\documentclass[conference]{IEEEtran}

\IEEEoverridecommandlockouts                              

\usepackage{helvet}         		
\usepackage{type1cm}        		
\usepackage{graphics} 		
\usepackage{graphicx}        
\usepackage[export]{adjustbox}
\usepackage{epsfig} 			
\usepackage{multicol}        		
\usepackage[hang,flushmargin]{footmisc}
\usepackage{amsmath}
\usepackage{amssymb}
\usepackage[ruled]{algorithm2e}
\usepackage{bm}
\usepackage{newtxtext}       %
\usepackage{newtxmath}       
\usepackage{caption}
\usepackage{gensymb}
\usepackage{graphicx}
\usepackage{booktabs}
\usepackage{xcolor}
\usepackage{epstopdf}
\usepackage{url}
\usepackage[font={small}]{caption}
\usepackage{commath}
\usepackage{subfigure}
\usepackage{multirow}
\usepackage{makecell}
\usepackage{calc}
\usepackage{pifont} 
\usepackage{tikz}
\usetikzlibrary{matrix}

\newcommand{\modified}[1]{\textcolor{black}{#1}}

\title{\vspace{5mm} \bf
\fontsize{16.1pt}{16pt}\selectfont SGM-SLAM: {\color{black}{
Scene Graph Matching for Data-Efficient Distributed SLAM}
}\vspace{-7mm}
}

\author{
Yewei Huang$^{1}$, Tixiao Shan$^{2}$, Abhinav Rajvanshi$^{2}$, Niluthpol Chowdhury Mithun$^{2}$, \\ Yaxuan Li$^{1}$, Brendan Englot$^{1}$ and Han-Pang Chiu$^{2}$
\vspace{-10mm}
\thanks{$^{1}$Y. Huang, Y. Li and B. Englot are with Stevens Institute of Technology, Hoboken, NJ, USA.
        {\tt\footnotesize \{yhuang85, yli21, benglot\}@stevens.edu}}
\thanks{Their work was supported in part by ONR grant N00014-24-1-2522.}
\thanks{$^{2}$T. Shan, A. Rajvanshi, N. Chowdhury and H. Chiu* are with SRI International, Princeton, NJ,  USA.
        {\tt\footnotesize \{tixiao.shan, abhinav.rajvanshi, niluthpol.mithun, han-pang.chiu\}@sri.com}}
\thanks{*The corresponding author is Han-Pang Chiu. \tt\footnotesize \ han-pang.chiu@sri.com}
}

\begin{document}
\maketitle 
\thispagestyle{empty}
\pagestyle{empty}
\begin{abstract}
\modified{We introduce a data-efficient distributed Simultaneous Localization and Mapping (SLAM) framework for multi-robot operation in GPS-denied and communication-constrained environments. The framework is designed for teams of robots equipped with LiDAR, cameras, and inertial sensors, where robots may operate from independently initialized local coordinate frames and share only limited spatial overlap.}
Our approach constructs a scene graph by 
using fused RGB-LiDAR point clouds to generate both a semantically segmented point cloud layer, and a layer of discrete bounded objects, to accompany estimated robot trajectories.
Scene graph matching is performed collaboratively through exchanging and matching object data with neighboring robots.
To maximize communication efficiency, we utilize a multi-step data exchange and optimization process. 
\modified{We demonstrate the effectiveness and communication efficiency of the proposed approach using real-world datasets containing limited trajectory overlap, viewpoint variation, repeated structures, and degraded visual conditions.}
\end{abstract}

\vspace{-4mm}
\section{Introduction}

\modified{Multi-robot Simultaneous Localization and Mapping (SLAM) is important for robotic teams operating in large-scale, GPS-denied environments. In applications such as emergency response, infrastructure inspection, and exploration, robots may enter from different locations, share limited trajectory overlap, and communicate over bandwidth-constrained networks. These conditions make inter-robot localization challenging, as feature-based approaches can be sensitive to viewpoint changes and perceptual degradation, while dense sensor-data exchange can impose substantial communication overhead. Robust inter-robot data association and communication-efficient information exchange are therefore essential for distributed multi-robot SLAM.}

\modified{Beyond accurate localization, field robots also benefit from semantic representations of their surroundings. Conventional point-cloud maps provide rich geometric information but can be expensive to communicate and difficult to interpret at the system or operator level. Semantic maps instead associate geometric structure with recognizable environmental entities, supporting higher-level reasoning, mission planning, and human interpretation. However, existing semantic SLAM methods~\cite{dube2020segmap, bowman2017probabilistic} often represent objects only as landmarks and provide limited relational structure between environmental entities.}

\begin{figure}[ht]
  \centering
    \includegraphics[width=\columnwidth]{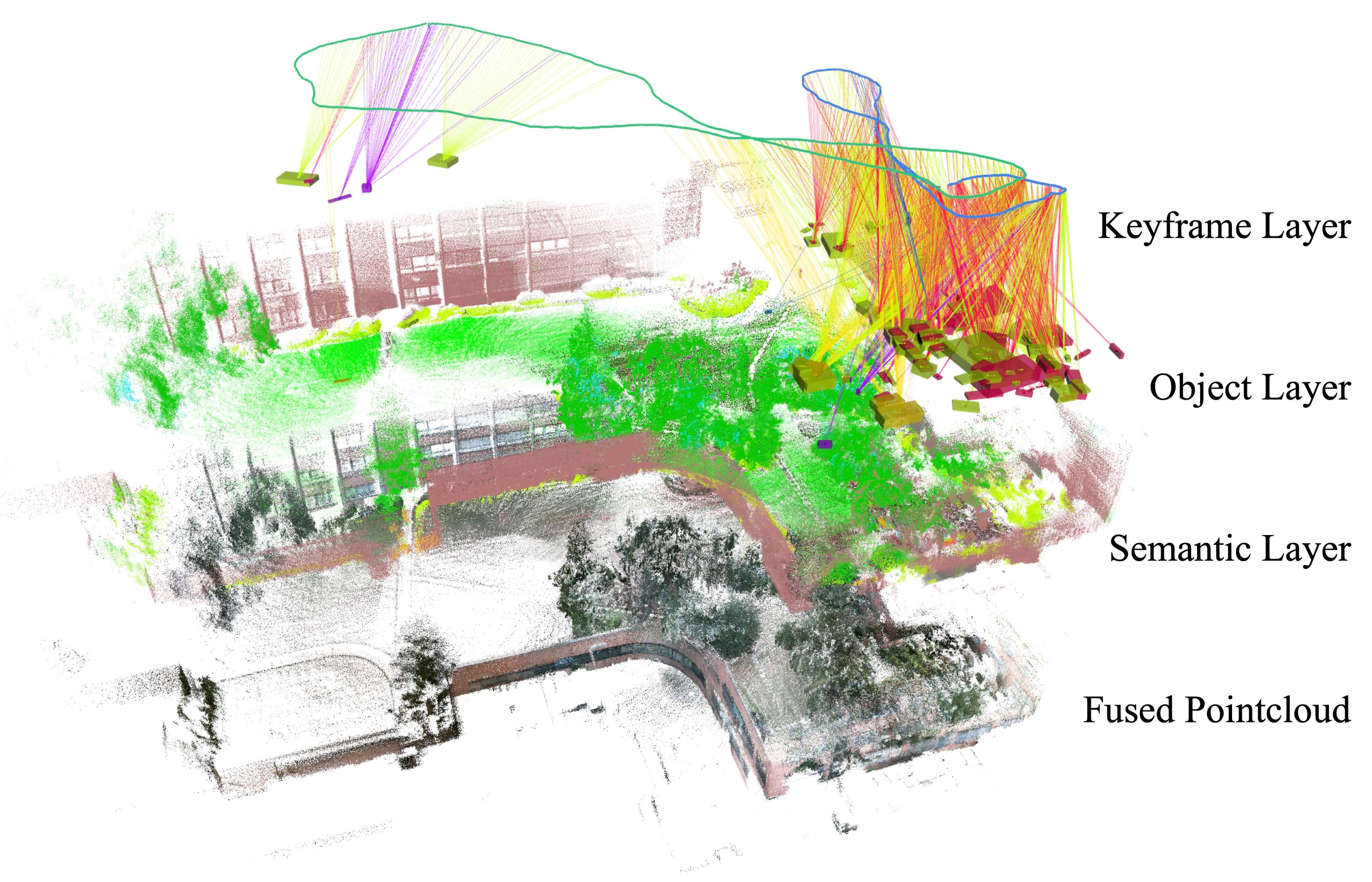}%
  \caption{\textbf{The merged result of an outdoor two-robot dataset 
  using scene graph matching.} The fused point-cloud data is from LiDAR and camera sensors. The trajectories of the two robots are represented in blue and green within the keyframe layer. Objects are represented as bounding boxes in various colors to indicate different object types in the object layer. In the semantic layer, the segmented point cloud is colorized by semantic classes.}\vspace{-6mm}
\label{fig:map_scene}%
\end{figure}

To achieve a better semantic understanding of the environment, researchers are now exploring the utilization of scene graphs in SLAM. 
A scene graph is a data structure that uses nodes in a graph to represent entities in the environment and edges to represent the relationships among these entities \cite{li2024scene}. 
As shown in Figure \ref{fig:map_scene}, a scene graph consists of entities such as objects, keyframes, and the associated semantic point clouds, with all their relationships represented as edges.
Compared to point cloud maps and semantic maps, scene graphs 
are easier to manage and their straightforward representation is intuitive for humans to understand and interact with.

Many SLAM methods with scene graphs have been developed for robot teams with cameras and LiDAR. 
However, most of these methods are either designed for indoor environments \cite{chang2023hydra, fernandez2024multi, gu2025mr} or 
specifically developed for outdoor autonomous driving environments \cite{greve2024collaborative, steinke2025collaborative}.
In addition, these methods typically rely on image-based descriptors for inter-robot data association.
We refer to this inter-robot data association process in scene graph SLAM as \textbf{scene graph matching} in this paper.
These high-dimensional descriptors may create a communication burden on the system, leading to map discrepancies and delayed updates, particularly in low-bandwidth outdoor environments.
A more communication efficient scene graph matching technique is needed for multi-robot SLAM systems working on long term missions in large-scale indoor and outdoor environments.

To address these issues, we propose SGM-SLAM, which performs scene graph matching efficiently using limited semantic and geometric data in distributed SLAM systems. 
SGM-SLAM is designed using a set of onboard sensors (LiDAR, camera, and inertial sensors) on each robot, without relying on external prior geometric information from GPS. 
In this algorithm, 
scene graph matching is initiated by matching the object data of the scene graph, 
making the system more robust to viewpoint changes across different robots.

Our contributions are summarized as follows:
\begin{itemize}
    \item We present and demonstrate a distributed multi-robot SLAM framework that constructs scene graphs for both indoor and outdoor environments, which detects scene graph matching candidates using only object semantics and geometry.
    \item A novel data-efficient communication strategy is introduced, which is designed to support distributed SLAM systems on robots equipped with LiDARs and cameras.
\end{itemize}

The remainder of the paper are organized as follows. 
Sec.~\ref{sec:Related} provides a brief overview of scene graphs and multi-robot SLAM algorithms.
Sec.~\ref{sec:Formulation} formulates the multi-robot SLAM problem in a distributed manner.
Sec.~\ref{sec:Algorithm} presents a detailed explanation of the proposed SGM-SLAM algorithm.
Sec.~\ref{sec:exp} demonstrates the performance of SGM-SLAM, and Sec.~\ref{sec:conclusion} concludes the paper.
\vspace{-2mm}
\section{Related Works}\label{sec:Related}

A key challenge that distinguishes \textbf{multi-robot SLAM} from single-robot SLAM is \textbf{inter-robot data association}. 
Unlike single-robot SLAM, which relies on the robot’s trajectory history to reject ambiguous loop closures, multi-robot systems lack such prior pose information for outlier rejection and point cloud registration. 
Limited communication bandwidth further requires selective data exchange, making efficient centralized or distributed strategies essential for robust inter-robot data association.
To address the data size issue for transmission, LiDAR-SLAM methods such as SWARM-SLAM~\cite{lajoie2023swarm} use compact descriptors to represent LiDAR scans.
LAMP~\cite{ebadi2020lamp} performs multi-sensor fusion and uses visual feature descriptors to provide candidates for LiDAR scan registration. While compact descriptors are relatively communication efficient, they often fail under viewpoint changes. To mitigate this, Zhou et al.~\cite{zhou2006multi} and CoLRIO~\cite{zhong2024colrio} restrict loop closure detection to rendezvous scenarios. However, this approach requires robots stay in close proximity, significantly limiting its applicability.

Some approaches attempt to address these issues. For example, 
Dubé et al.~\cite{dube2017online} introduce semantic segmentation to represent environments. However, these segmented maps are still represented as point clouds, which can impose a communication burden for large-scale environments. 
MG-GMMapping~\cite{dong2022mr} represents the environment using a Gaussian Mixture Model, yet inter-robot loop closure candidates are still determined by compact descriptor matching.

\textbf{Scene graphs,} initially introduced by the computer vision community as a semantic image retrieval technique~\cite{johnson2015image}, have gained increasing attention in robotics due to their potential to support human-robot interaction~\cite{rosinol2021kimera}.
The scene graph commonly adopted in robotics is a 3D scene graph~\cite{armeni20193d}, which uses a graph-based structure to represent the semantic information of a 3D model of the environment. Unlike semantic maps~\cite{dube2020segmap, bowman2017probabilistic}, a scene graph not only encapsulates the semantic details of the environment but also explicitly describes the relationships between semantic entities.

Most multi-robot SLAM works with scene graphs use visual descriptors for scene graph matching. 
Hydra~\cite{hughes2024foundations} and Hydra-multi~\cite{chang2023hydra} introduce hierarchical descriptors, with the lowest level based on bag-of-words descriptors. 
Moreover, Hydra-multi operates as a centralized system, with a central node responsible for all scene graph construction and matching.
While Kimera-Multi~\cite{tian2022kimera} is a distributed system, its scene graph matching also relies on visual bag-of-words descriptors.
Radwan et al.~\cite{radwan2024uav} introduce visual fiducial markers to enable scene graph construction in under-construction environments.
MR-COGraphs~\cite{gu2025mr} proposes a learning-based, data-efficient visual feature method for scene graph matching. The only exception is Multi S-graphs~\cite{fernandez2024multi}, which includes LiDAR in the multi-robot SLAM system. However, they use specific room-based compact descriptors for scene graph matching, which only work in indoor environments.
There are also LiDAR-based SLAM works with scene graphs in outdoor environments \cite{greve2024collaborative}, \cite{deng2024opengraph}. 
However, in these works, the scene graph serves only as a product, and its structure does not directly contribute to robot localization. 
CURB-OSG \cite{steinke2025collaborative} is the first to perform scene graph matching with unknown initial alignment in outdoor environments.
However, since it is a centralized method, the system tends to suffer from delayed map updates due to limited communication bandwidth.

Many robust \textbf{object-level loop closure detection} methods have been proposed for object-based SLAM. 
Qin et al.~\cite{qin2021semantic} and SemanticLoop~\cite{yu2022semanticloop} formulate object-level loop closure as a bipartite graph matching problem, using both semantic and geometric information of the objects.
In addition to semantic and geometric information, SymbioLCD2 and Ji et al. \cite{ji2023loop} add visual learning-based descriptors for object data association.
Liu et al. \cite{lin2021topology} combine semantic and geometric information with the corresponding image histograms. 
To further enhance robustness, SemanticTopoLoop~\cite{cao2024semantictopoloop} introduces a multi-layer object data association strategy that considers semantic and geometric information alongside probabilistic criteria. 
However, since these methods are primarily designed for single-robot SLAM problems, they do not consider the communication bandwidth limitations in multi-robot scenarios.

Inspired by these object-based loop closure detection methods for single robot SLAM, we propose SGM-SLAM, a robust and communication-efficient distributed SLAM method that works for low-bandwidth indoor and outdoor environments using limited object data. 
The scene graph from each robot is constructed using keyframe, object and point cloud information from a semantic SLAM system with visual, LiDAR, and inertial sensors. 
Our approach reduces data transmission by exchanging only the semantic labels and centroids of grouped objects across robots, and performing object matching to search for scene graph matching candidates. 
The estimated transformations between grouped objects serve as geometric priors for inter-robot point cloud registration, helping to reduce failures caused by outliers.
\vspace{-6mm}
\section{Problem Formulation and Approach}\label{sec:Formulation} \vspace{-1mm}
The key problem in multi-robot SLAM is determining the transformation from each robot's local frame to a common global frame.
We aim for a simplified approach that achieves promising performance even under limited communication bandwidth, while ensuring consistent and reliable robot state estimation.
This paper adopts a two-fold approach: robot frame transformation estimation followed by local SLAM pose graph optimization \cite{factor_graphs_for_robot_perception} (PGO) with relevant neighboring robot states.

We define the set of $n$ robots in the team as $\mathcal{R} = \{R_0, \dots, R_{n-1}\}$.
For each robot $R_{\alpha} \in \mathcal{R}$, we align the local world frame $\mathbf{W}_\alpha$ with $\mathbf{B}_{\alpha_0}$, the body frame $\mathbf{B}$ of the local robot $\alpha$ at the initial timestep $t_0$.
This allows us to transform the inter-robot data association problem into estimating the relative transformations $\mathcal{T} = \{\mathbf{T}_{\mathbf{W}_\alpha \rightarrow \mathbf{W}_\beta}\mid R_\beta \in \mathcal{R}\}$ 
from the local world frames of other robots $\mathbf{W}_\beta$ to the local world frame $\mathbf{W}_\alpha$.

The state (a 6 DoF pose) of robot $R_{\alpha}$ at a timestep $t_i$ is defined as $\mathbf{x}_{\alpha_i}$.
The complete trajectory of robot $R_\alpha$ from the initial timestep $t_0$ to the current timestep $t_i$ is denoted as $\mathcal{X}_{\alpha} = \{ \mathbf{x}_{\alpha_0}, \dots, \mathbf{x}_{\alpha_i}\}$.
We consider three types of observations in this paper:
\begin {itemize}
    \item \textbf{odometry} observation $\mathcal{Z}_{\alpha}^{\text{odom}} = \{ \mathbf{z}_{\alpha_i,\alpha_{i+1}}^{\text{odom}} \}$, for sequential matching;
    \item \textbf{loop closure} observation $\mathcal{Z}_{\alpha}^{\text{intra}} = \{ \mathbf{z}_{\alpha_i,\alpha_{j}}^{\text{intra}} \}$, for intra-robot data association within the local robot trajectory;
    \item \textbf{inter-robot} observation $\mathcal{Z}_{\alpha}^{\text{inter}} = \{ \mathbf{z}_{\alpha_i,\beta_j}^{\text{inter}} \mid \beta \in \mathcal{R} ,\beta \neq \alpha \}$, for data association with inter-robot point-cloud registration between the local and neighboring robot trajectories.
\end {itemize}
To simplify the approach, we do not share our detected inter-robot observations with neighboring robots. 
Therefore, $\mathbf{z}_{\beta_i,\alpha_j}^{\text{inter}}$ is only considered an observation for robot $\beta$, not for robot $\alpha$.
For robot $\alpha$, the complete observation set is $\mathcal{Z}_{\alpha} = \{ \mathcal{Z}_{\alpha}^{\text{odom}} \cup \mathcal{Z}_{\alpha}^{\text{intra}} \cup \mathcal{Z}_{\alpha}^{\text{inter}}\}$.

In the global step, we start by constructing fully connected graphs $\mathcal{G}_{\alpha}$ and $\mathcal{G}_{\beta}$ using selected semantic objects from the local scene graph and the objects received from the neighboring robot $R_\beta$.
These object graphs are then matched using a graph matching algorithm similar to SemanticLoop \cite{yu2022semanticloop}.
Once the overlapping component $\mathcal{G}_{\alpha} \cap \mathcal{G}_{\beta}$ is identified, the relative transformation $\mathbf{T}_{\mathbf{W}_\alpha \rightarrow \mathbf{W}_\beta}$ can be estimated through rigid transformation calculations based on matched object positions, and further refined using associated object point clouds. A detailed explanation of the relative transformation estimation can be found in Sec. \ref{sec:Transformation}.

In the local step, each robot performs PGO. The  problem is formulated as a maximum a posteriori estimation problem~\cite{5979641}:
\begin{align}\label{eq:slam_eqn}
    {\mathcal{X}_\alpha}^*, {\mathcal{X}_{\text{assoc}}}^* &= \mathop{\arg\max}_{\mathcal{X}_\alpha, \mathcal{X}_{\text{assoc}}} P(\mathcal{X}_\alpha, \mathcal{X}_{\text{assoc}} \mid \mathcal{Z}_\alpha),
\end{align}
where $\mathcal{X}_{\text{assoc}} = \{ \mathbf{x}_{\beta_j} : \mathbf{z}^\text{inter}_{\alpha_i, \beta_j} \in \mathcal{Z}^\text{inter}_\alpha\}$
denotes the states of neighboring robots that are associated with the local robot due to detected inter-robot observations.

\section{Algorithm}\label{sec:Algorithm} 
\begin{figure}[t]
  \centering
    \includegraphics[width=.9\columnwidth]{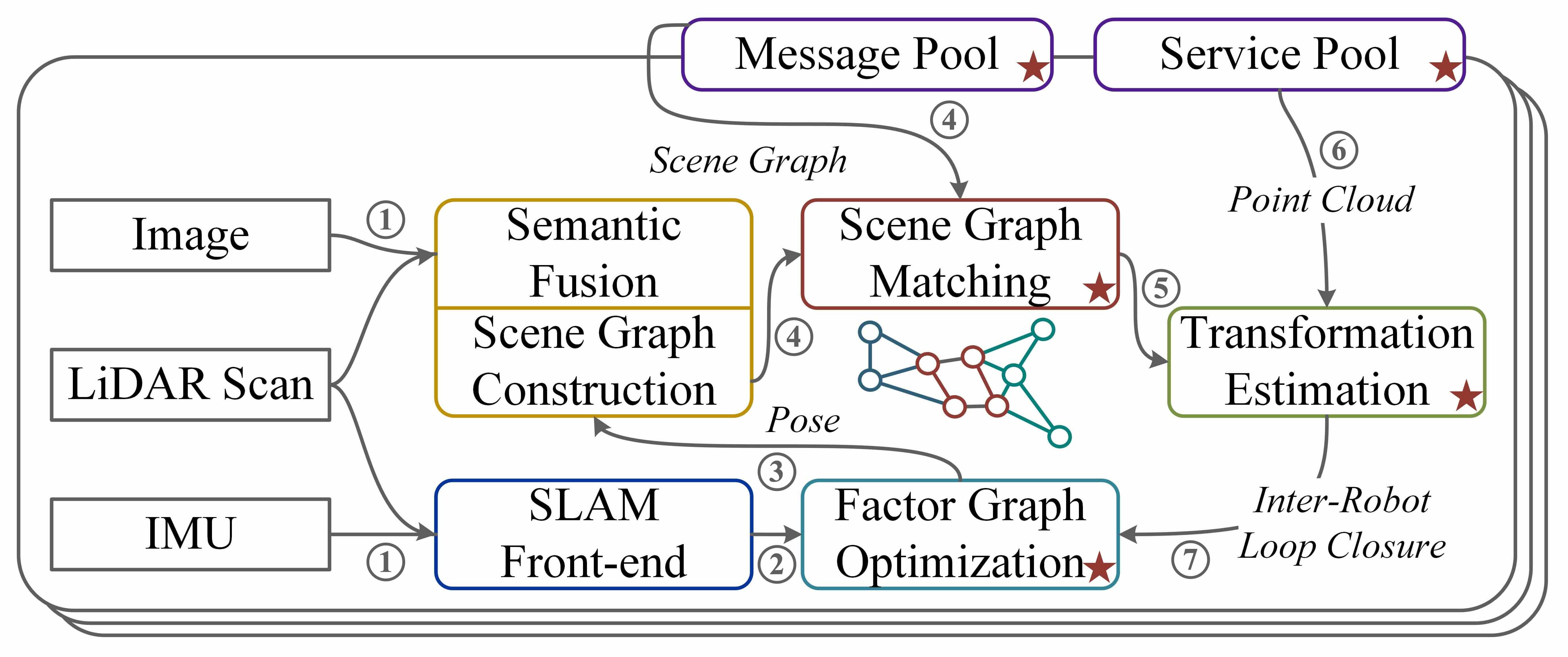}%
  \caption{\textbf{System Architecture.} The SGM-SLAM pipeline for each robot, with inter-robot parts marked with red stars.}\vspace{-4mm}
\label{fig:pipeline}%
\end{figure}
In this section, we present our multi-robot LiDAR-Visual-Inertial SLAM algorithm, and describe the scene graph matching process. The system architecture within each robot is illustrated in Fig. \ref{fig:pipeline}.
We perform semantic sensor fusion using both camera and LiDAR data, while simultaneously estimating the robot states through LiDAR-Inertial odometry and intra-robot loop-closures. 
We employ Point-LIO \cite{he2023point} to compute LiDAR-Inertial odometry, and intra-robot loop closures are detected using a Euclidean distance-based approach similar to that of LIO-SAM \cite{shan2020lio}.

Next, we construct a scene graph based on the estimated robot states and the semantically labeled point-cloud from the semantic fusion. 
The object data (semantic labels and centroids) of this scene graph is then shared with robot neighbors. 
Once the local robot receives the object data from a robot neighbor, scene graph matching is performed. 
If a sufficient number of objects in the scene graph are matched, a relative transformation is computed to determine the inter-robot loop closures.
Finally, all the observations are added to the SLAM factor graph. 
Both the local robot states and the scene graph are then subsequently updated.

\begin{figure}
  \centering
    \subfigure{
        \includegraphics[width=.47\columnwidth]{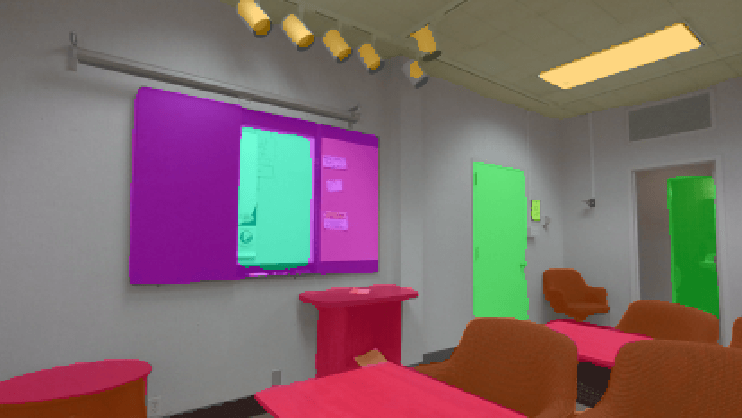}
    } 
    \subfigure{
        \includegraphics[width=.47\columnwidth]{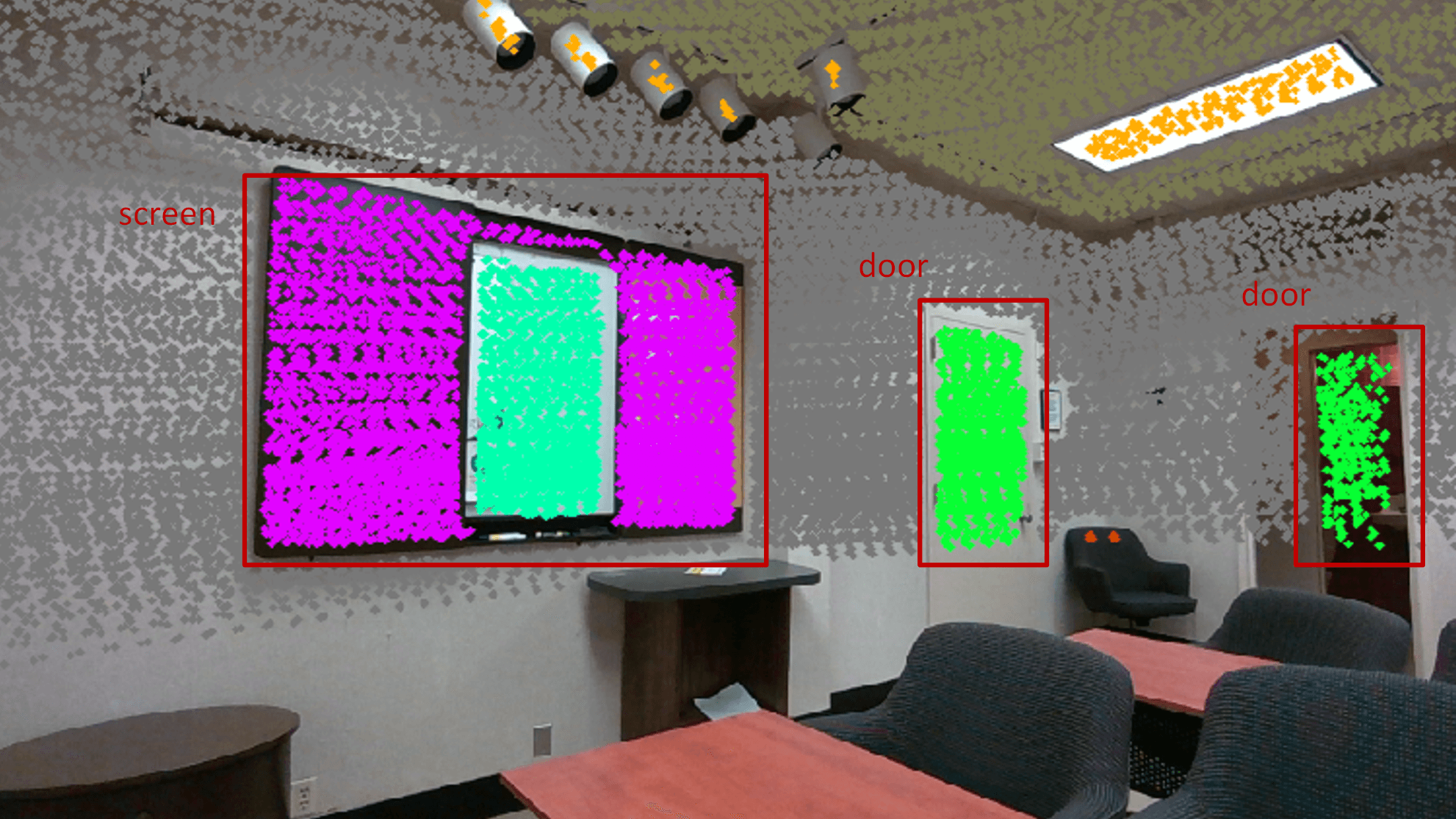}
    }
    \subfigure{
        \includegraphics[width=.65\columnwidth]{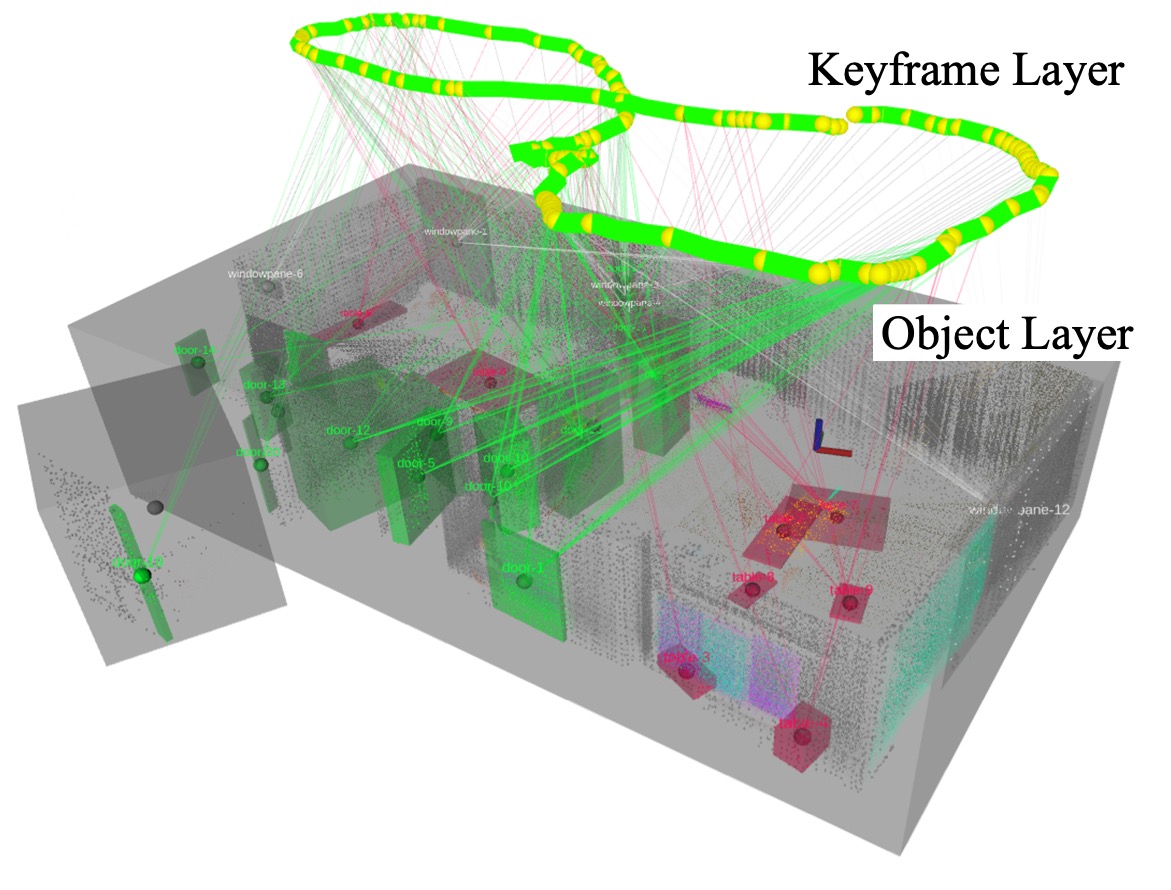}
    }
    \vspace{-2mm}
  \caption{Segmented image with detected objects shown in different colors (top left) and LiDAR scan projected onto the segmented image, with objects of interest marked by red rectangles (top right). The bottom shows an example of the connectivity between the keyframe layer and the object layer.} \label{fig:segmentation}\vspace{-5mm}
\end{figure}

\subsection{Scene Graph Construction}
\vspace{-1mm}
Fig. \ref{fig:map_scene} illustrates a scene graph constructed by two robots.
The keyframe layer includes keyframes selected by the SLAM front-end as vertices, connected by odometry observations between them, along with loop closure observations and inter-robot observations. 
The object layer is a fully connected graph where object vertices are interconnected by edges, using Euclidean distances between objects as the edge weights. 
The semantic layer contains the labeled semantic point clouds.
Cross-layer connections link keyframe vertices to object vertices when the corresponding object is observed in that keyframe. 
Additionally, each scan in the semantic layer is linked to the keyframe where it was captured.

The object vertices in the object layer are detected through semantic sensor fusion. During this process, RGB images are aligned with LiDAR scans to create a semantically labeled point cloud containing objects of interest. 
As shown in the top part of Fig. \ref{fig:segmentation}, the RGB images are segmented using Oneformer~\cite{jain2023oneformer}. 
We perform segmentation using only image data to ensure the consistency and accuracy of the semantic labeling.
The point set from the corresponding LiDAR scan is then transformed into the 2D RGB image frame utilizing the LiDAR-camera extrinsics and camera intrinsics.
Subsequently, the semantic label from the nearest pixel is assigned to the corresponding point.
Finally, the labeled point cloud is clustered into distinct objects.

The object layer is constructed using objects (nodes) across keyframes, with relationships (edges) weighted by Euclidean distances among objects. 
It can be formulated as a Linear Assignment Problem (LAP):
\begin{align}\label{eq:lap}
    A_n^* &= \arg \min_{A_n}\sum C \odot A_n\\
     &= \arg \min_{A_n} \sum_{v_i \in \mathcal{V}_{\alpha}^{(t)}} \sum_{v_j \in \mathcal{V}_{\alpha}^{(t-1)}} c_{ij} \cdot a_{ij}; \\
     \text{s.t. } 
    &\forall v_j \in \mathcal{V}_{\alpha}^{(t-1)}, \sum_{v_i \in \mathcal{V}_{\alpha}^{(t)}} a_{ij} \leq 1; \nonumber\\
    &\forall v_i \in \mathcal{V}_{\alpha}^{(t)}, \sum_{v_j \in \mathcal{V}_{\alpha}^{(t-1)}} a_{ij} \leq 1. 
\end{align}
Given two sets of object candidates (vertices) at the current keyframe timestep $t$ and the previous keyframe timestep $t-1$, denoted as $\mathcal{V}_{\alpha}^{(t)}$ and $\mathcal{V}_{\alpha}^{(t-1)}$, we define the cost matrix $C$ such that each element $c_{ij} = \norm{p_i - p_j}_2$ represents the Euclidean distance between the object centroids $p_i$, $p_j$ of the two vertices.
$A_n$ is the Boolean assignment matrix indicating object vertex matches between two sets with elements $a_{ij} \in \{0,1\}$. If $a_{ij} = 1$ and the cost $c_{ij} < c_{\max}$, then the $i_\textsuperscript{th}$ vertex $v_{i}\in \mathcal{V}_{\alpha}^{(t)}$ and $j_\textsuperscript{th}$ vertex $v_{j}\in \mathcal{V}_{\alpha}^{(t-1)}$ are matched.
$c_{\max}$ represents the maximum allowable distance for two object candidates across keyframes to be identified as the same object.
Objects are allowed to remain unmatched to accommodate newly observed objects and objects that are no longer tracked.
We use the Hungarian Algorithm~\cite{kuhn1955hungarian} to solve this LAP.
One benefit of using a hierarchical graph structure is the convenience of propagating updates across layers.
The bottom part of Fig.~\ref{fig:segmentation} illustrates the connections between keyframes and objects.
While a loop closure happens and the keyframe poses are adjusted, object positions are also sequentially updated based on the latest keyframe that observes each object.

\begin{figure}[t!]
  \centering
  \begin{subfigure}
    \centering
    \includegraphics[width=.7\columnwidth]{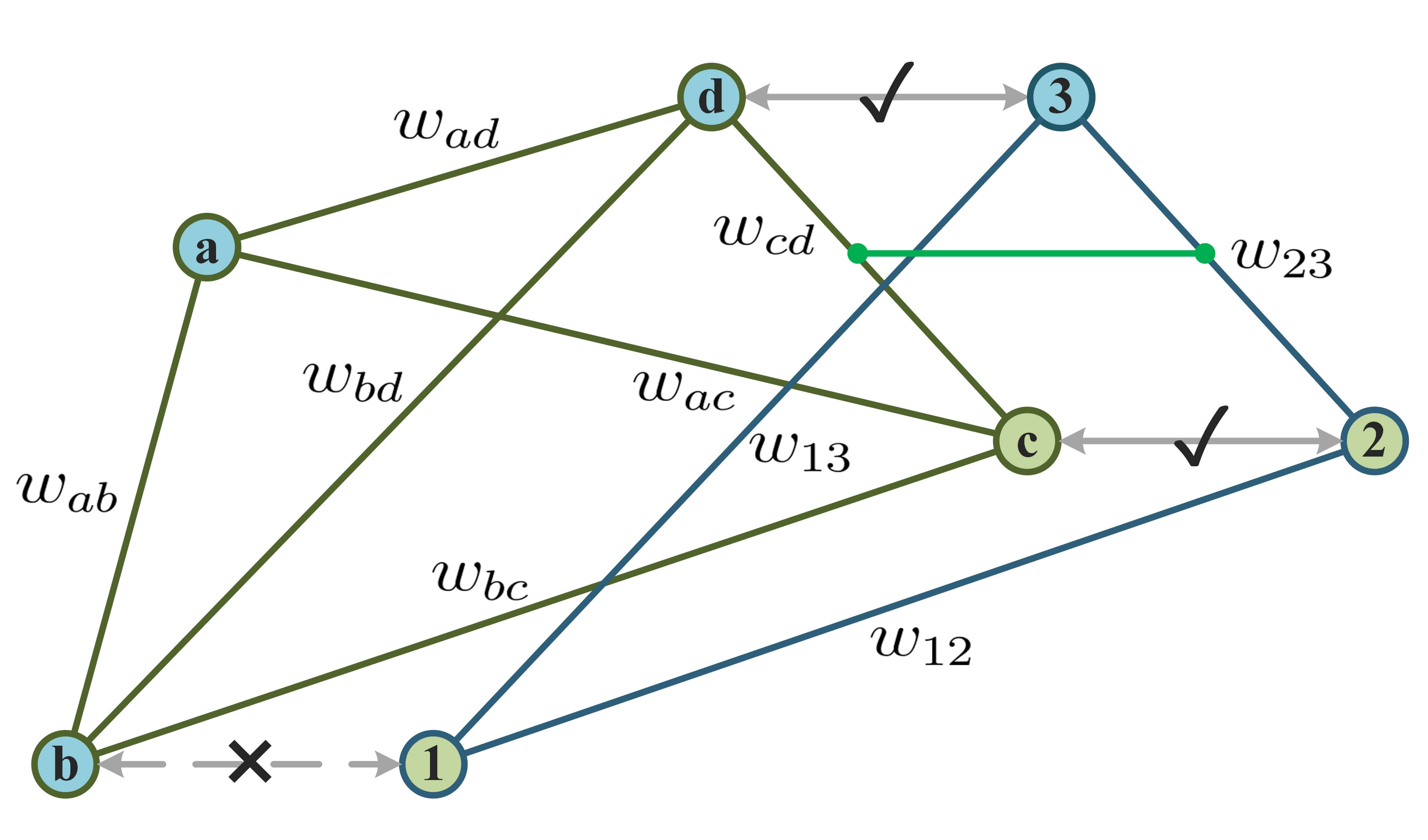}
    \label{fig:matching_a}
  \end{subfigure}
  \vspace{-2mm}
  \begin{subfigure}
    \centering
    \includegraphics[width=\columnwidth]{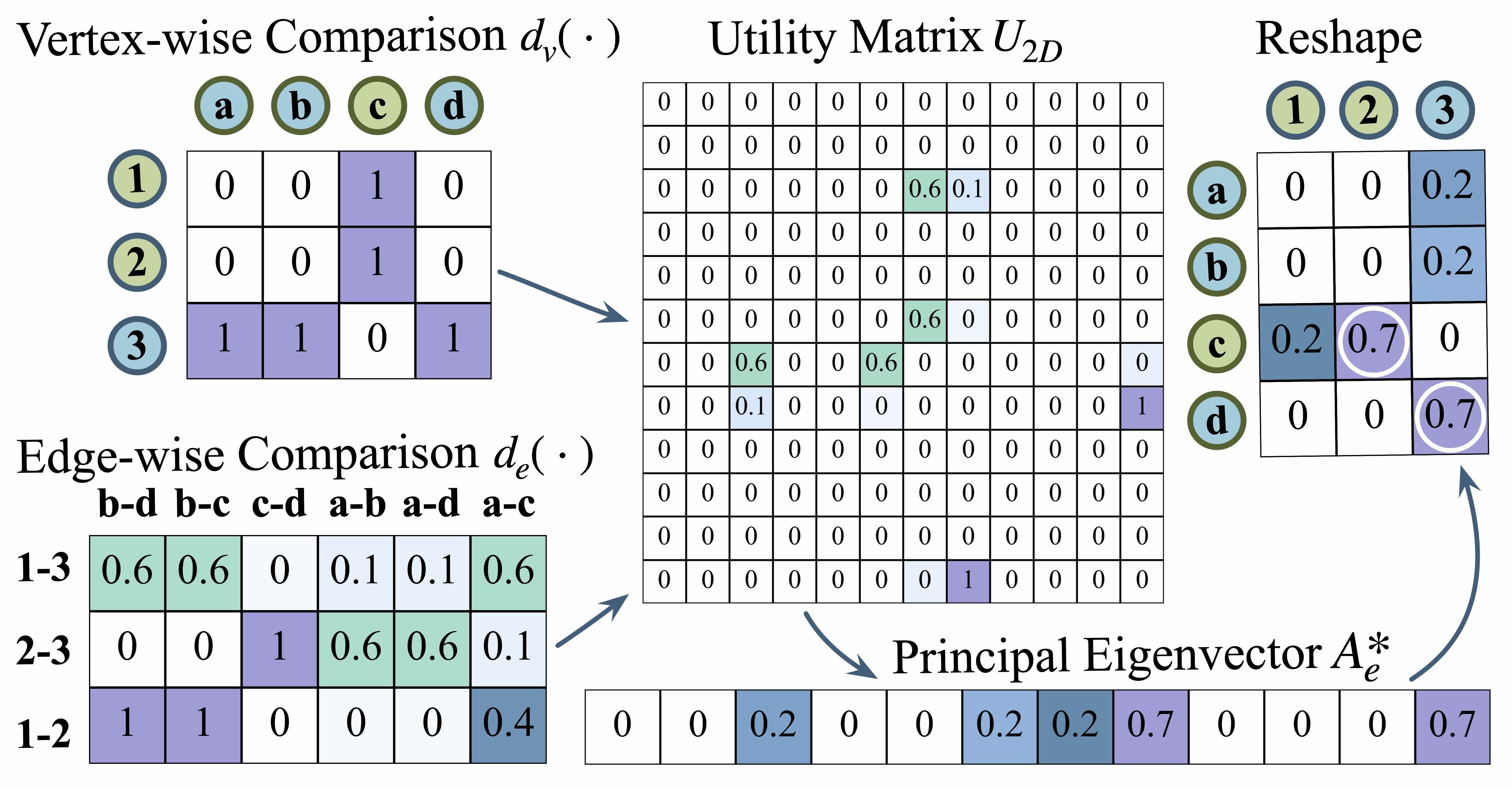}
    \label{fig:matching_b}
  \end{subfigure}
  \vspace{-4mm}
  \caption{An example illustrating edge-wise matching between two graphs. Edge lengths represent weights, and node colors (green and blue) indicate vertex types.}
  \label{fig:matching}
  \vspace{-5mm}
\end{figure}

\vspace{-2mm}
\subsection{Scene Graph Matching with Object Data}\label{sec:Scene-graph-matching}
\vspace{-2mm}
In the scene graph matching step, we present the object data as graphs and match them with graph matching.
During scene graph matching, object data is used to construct a fully-connected undirected graph. 
We use only selected object classes for matching since the estimated centroids of larger objects could vary significantly when observed from different viewpoints.
We define the object as a vertex $v_i$ with attribute $\phi(v_i) = (p_i, l_i)$, where $p_i$ denotes the centroid of the object and $l_i$ denotes its semantic label.
The edge $e_{ij}$ connects vertices $v_i$ and $v_j$, with weight $w_{ij} = \norm{p_i - p_j}_2$ being the Euclidean distance between the centroids of two objects.

Define the graph of the local robot $\alpha$ as $\mathcal{G}_{\alpha} = \langle \mathcal{V}_{\alpha}, \mathcal{E}_{\alpha} \rangle$, where $\mathcal{V}_{\alpha} = \{v_i\}$ denotes the set of vertices and $\mathcal{E}_{\alpha} = \{e_{ij}\}$ denotes the set of edges connecting them. 
The graph constructed using object data received from a neighbor robot $\beta$ is denoted as $ \mathcal{G}_{\beta} = \langle \mathcal{V}_{\beta}, \mathcal{E}_{\beta} \rangle$.
This graph matching problem can be viewed as a maximum bipartite matching problem. 
Since each robot has its own local world frame, we cannot directly use Euclidean distances between object centroids for graph matching, as is done in the intra-robot case (Sec.~\ref{sec:Algorithm}.A) when constructing the object layer.
Instead, we adopt a similar technique used in SemanticLoop \cite{yu2022semanticloop}, and formulate the problem as a Quadratic Assignment Problem (QAP):
\vspace{-2mm}
\begin{align}\label{eq:qap}
     A_e^*      &=\arg \max_{A_e} \sum A_e \odot U \odot A_e\\
     &=\arg \max_{A_e} \sum_{v_i, v_j \in \mathcal{V}_{\alpha}} \sum_{v_m, v_n \in \mathcal{V}_{\beta}} a_{im} \cdot u_{ijmn} \cdot a_{jn};  \\
     \text{s.t. } 
         &\forall v_i \in \mathcal{V}_\alpha, \sum_{v_m \in \mathcal{V}_\beta} a_{im} \leq 1; \nonumber \\
    &\forall v_m \in \mathcal{V}_\beta, \sum_{v_i \in \mathcal{V}_\alpha} a_{im} \leq 1.\label{eq:qap_constraint}
    \vspace{-2mm}
\end{align}
Here $A_e$ is the Boolean assignment matrix where each element $a_{im} \in \{0,1\}$ indicates the matching status of vertices $v_i \in \mathcal{V}_{\alpha}$ and $v_m \in \mathcal{V}_{\beta}$.
$a_{im} = 1$ means $v_i$ and $v_m$ are matched.
$u_{ijmn}$ is an element of the 4D utility tensor $U$, measuring the similarity between edge $e_{ij}$ from $\mathcal{G}_{\alpha}$ and edge $e_{mn}$ from $\mathcal{G}_{\beta}$:
\vspace{-2mm}
\begin{equation}
    u_{ijmn} = d_v(v_i, v_m) \cdot d_v(v_j, v_n) \cdot d_e(e_{ij}, e_{mn}).
    \vspace{-2mm}
\end{equation}
The function $d_v(\cdot)$ represents the vertex-wise comparison. 
In our case, we compare only the labels of two objects to avoid confusion caused by partially observed large objects.
If two objects have the same label, we set $d_v(\cdot) = 1$; otherwise, $d_v(\cdot) = 0$.
The edge-wise comparison is defined as  $d_e(e_{ij}, e_{mn}) = \exp(-\mu|w_{ij} - w_{mn}|)$, where $w_{ij}$ is the weight of edge $e_{ij}$ and $\mu$ is a scaling factor.
Fig. \ref{fig:matching} showcases a simple example of edge-wise matching with $\mu = 1$.
In this figure, the edge lengths represent the actual weights of the graph.
Consequently, $d_e(w_{13}, w_{bd})$, $d_e(w_{12}, w_{bc})$ and $d_e(w_{23}, w_{cd})$ are significantly higher than the $d_e(\cdot)$ values of other matches.
However, since vertex $b$ and vertex $1$ are of different types, $d_v(v_b, v_1) = 0$, resulting in only $e_{23}$ and $e_{cd}$ being matched.

We then vectorize the assignment matrix as $\text{vec}(A_e)$, and reshape $U$ into a 2D matrix $U_{\text{2D}}$, allowing for the rewriting of Eq. (\ref{eq:qap}):
\vspace{-2mm}
\begin{equation}
    A_e^* = \arg \max_{A_e} {\text{vec}(A_e)}^{\top} \cdot U_{\text{2D}} \cdot \text{vec}(A_e).
    \vspace{-2mm}
\end{equation}
In Eq. (\ref{eq:qap}), $a_{ij}$ is originally constrained to be Boolean. 
However, to make the problem more tractable, we relax this constraint by allowing $a_{ij} \in [0, 1]$. 
As proved by Leordeanu et al.~\cite{1544893}, the solution, $A_e^*$, is then given by the positive eigenvector of $U_{\text{2D}}$ corresponding to its principal eigenvalue.
Finally, the vertex-wise matching of the objects can once again be formulated as an LAP similar to Eq. (\ref{eq:lap}), with cost matrix $C = -A^*_e$.
Additionally, a minimum allowable eigenvector value is introduced to prevent mismatches during the assignment.
\vspace{-2mm}
\subsection{Transformation Estimation}\label{sec:Transformation}
The transformation estimation module provides an initial guess for inter-robot point-cloud registration.
Let the matched vertex pairs from local graph $\mathcal{G}_\alpha$ and neighbor graph $\mathcal{G}_\beta$ be denoted by $\mathcal{A}_{\alpha,\beta}$. 
Our goal is to estimate the relative robot frame transformation $\mathbf{T}_{\mathbf{W}_\alpha \rightarrow \mathbf{W}_\beta}$ using this matched pair set alone.
For each matched object pair $\forall \langle v_{\alpha}, v_{\beta}\rangle \in \mathcal A_{\alpha,\beta}$, $p_{\alpha}$ and $p_{\beta}$ represent the object centroids of the corresponding vertices.
The transformation relationship between these points can be expressed as:
\vspace{-2mm}
\begin{equation}
    [\begin{matrix} p_{\alpha} \\ 1\end{matrix}] = \mathbf{T}_{\mathbf{W}_\alpha \rightarrow \mathbf{W}_\beta} [\begin{matrix} p_{\beta} \\ 1\end{matrix}].
    \vspace{-2mm}
\end{equation}
As long as we have more than 3 non-collinear objects, we can estimate $\mathbf{T}_{\mathbf{W}_\alpha \rightarrow \mathbf{W}_\beta}$ using a rigid body transformation. 
To avoid potential tilting due to odometry drift and ensure robustness, we assume that all objects lie on a plane, allowing us to reduce the problem to a 2D rigid body transformation estimation. 
\vspace{-2mm}
\subsection{Inter-robot Point-Cloud Registration}\label{sec:inter-robot-loop}
The robot frame transformation estimate $\mathbf{T}_{\mathbf{W}_\alpha \rightarrow \mathbf{W}_\beta}$ is used as the coarse initial transformation guess for inter-robot point-cloud registration.
We then refine the result by performing Iterative Closest Point (ICP) registration using the sparse object cloud associated with the matched objects.

In the scene graph, each object node $v$ is connected to a set of keyframes (point clouds), $\mathcal{F}_v = \{F(i)\}$ in which the corresponding object is observed.
For each matched object pair $\langle v_{\alpha}, v_{\beta}\rangle$, point-cloud registration is performed as follows.
The target point cloud for registration is constructed by aggregating all keyframes associated with the object from local scene graph: $M_{v_\alpha} = \bigcup_i F_{v_\alpha}(i)$.
Each $F_{v_\beta}(j) \in \mathcal{F}_{v_\beta}$ is processed using a sliding-window approach to construct a source cloud, $M_{v_\beta} = \cup^{5}_{k=-5} F_{v_\beta}(j+k)$.
This source cloud $M_{v_\beta}$ is then transformed into local world frame $\mathbf W_{\alpha}$ using the transformation $\mathbf{T}_{\mathbf{W}_\alpha \rightarrow \mathbf{W}_\beta}$.
Point cloud registration between each source cloud and the target cloud is performed using ICP, and the resulting inter-robot loop closures are added into the SLAM factor graph to improve local robot state estimation.
To ensure computational efficiency, each keyframe from the neighboring robot is registered with only one local keyframe.

\subsection{Communication Between Robots}
The robots continuously share the object data with neighboring robots.
Only the object ID, type and centroid position are shared with the neighbors at a relatively high frequency.
After receiving object data from a neighbor, the robot stores it and compares it with its local object data.
If a sufficient number of matched objects are detected, the robot sends a service request to the corresponding neighbor for the sparse object point cloud and the associated keyframe IDs of the matched objects.
Then, the local robot checks the neighbor's keyframe database, and sends a service request for the compressed point cloud, optimized robot state estimate, and marginalized covariances of any keyframes it needs but has not yet stored.
To optimize communication bandwidth, the shared point cloud includes only geometric information (pose); intensity and color data are not transmitted between neighbors.
The point cloud is compressed using the Point Cloud Library (PCL) \cite{Rusu_ICRA2011_PCL}.
Upon receiving the request, the neighbor robot searches its dataset and sends the required data.  
This data-efficient message-passing strategy ensures that the latest lightweight object data is continuously shared, while the bandwidth-intensive compressed point cloud is transmitted only when required.

\begin{figure}[t]
  \centering
    \includegraphics[width=.65\columnwidth]{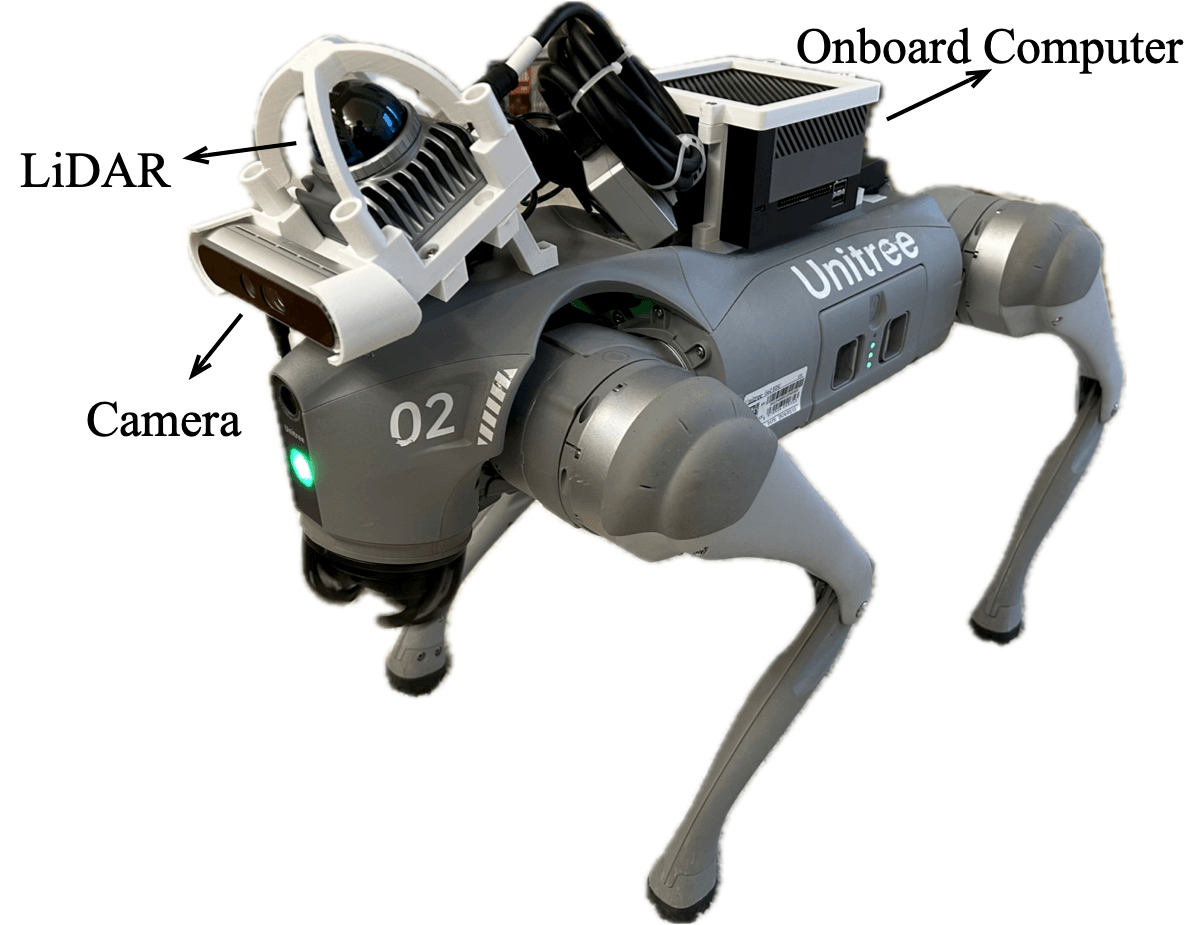}%
  \caption{Our data-gathering platform consists of a Unitree quadrupedal robot with sensors mounted on the top.}\vspace{-5mm}
\label{fig:sensor}%
\end{figure}
\vspace{-2mm}
\section{Experiments}\label{sec:exp}
We evaluate the performance of our proposed scene graph matching with object data algorithm using both simulation and real-world datasets.
The simulation dataset is used to test the algorithm's capability in performing sufficient graph matching and transformation estimation.
The real-world datasets in this paper were collected using a heterogeneous multi-robot setup, which includes a handheld device and a Unitree Go2 EDU, as shown in Fig.~\ref{fig:sensor}. Both platforms are equipped with a Livox mid-360 LiDAR, an inertial sensor, and an Intel RealSense Depth Camera. We note that only the RGB image from this sensor is utilized in our work. 
Furthermore, the onboard sensors of the Unitree quadrupedal robot are not utilized in this work; instead, the same handheld device is attached to the quadrupedal robot for data collection. 

We collected this dataset specifically because existing multi-robot SLAM datasets~\cite{zhu2023graco, tian2023resilient, feng2024s3e} use a Velodyne VLP-16 LiDAR, which produces sparse, layered point clouds within a fixed vertical field of view.
The low vertical resolution and the large inter-layer gaps make it difficult to directly apply the clustering-based algorithm we used for object detection on the denser Livox point-cloud.
The data sequences are collected in a suburban campus environment.
Due to limitations in equipment availability, we collected sequences for each robot individually as ROS2 bag files and later synchronized them offline by playing them simultaneously.
The entire system is implemented using Robot Operating System (ROS) 2 Foxy on Ubuntu 20.04. 
The GPU is utilized only for image segmentation.

\begin{figure}[htp]
  \centering
    \includegraphics[width=.95\columnwidth]{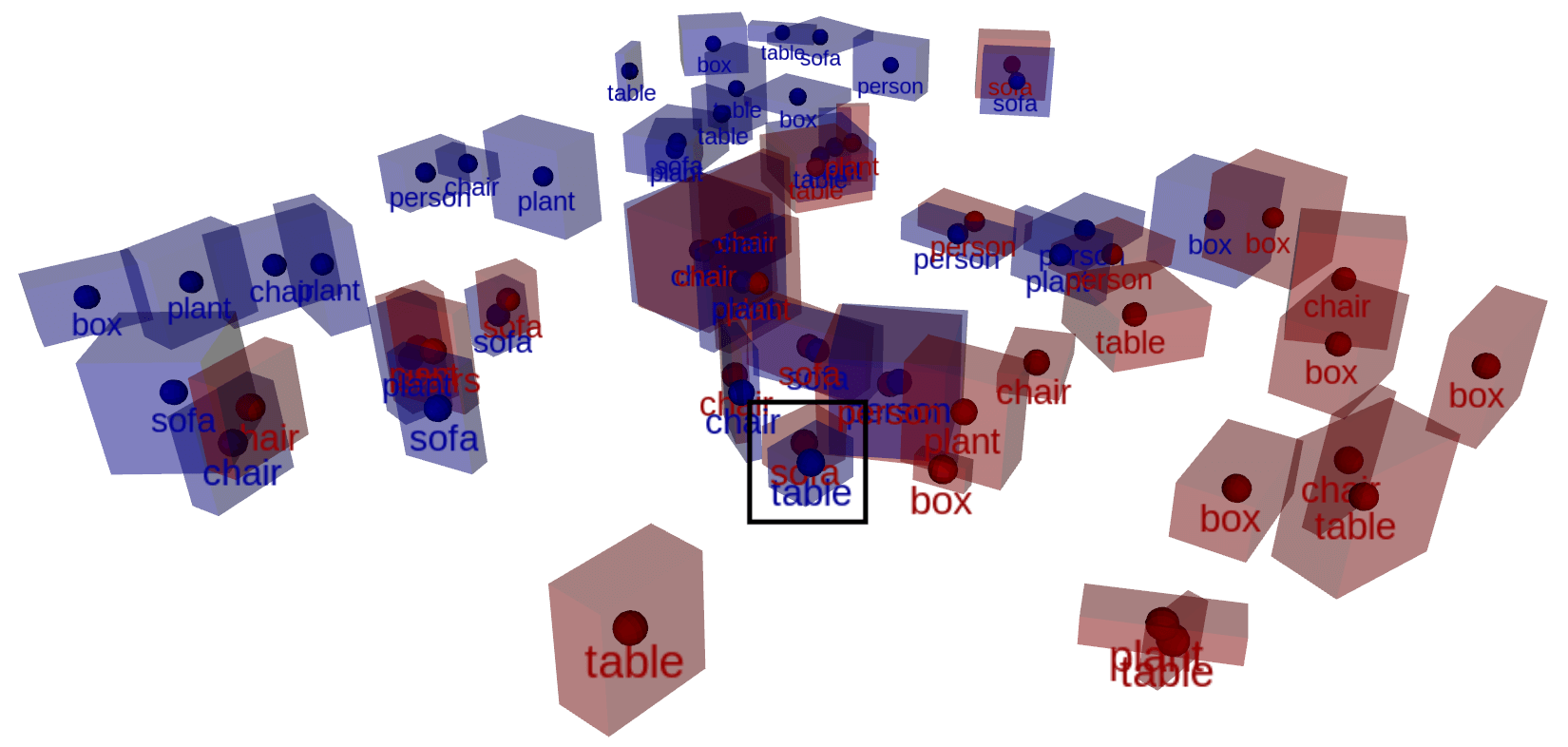}%
  \caption{An example illustrating a successful transformation estimation despite the presence of both position and semantic label errors. Objects in $\mathcal{G}_1$ are marked with red cuboids, and objects in $\mathcal{G}_2$ are marked with blue cuboids. The objects in $\mathcal{G}_2$ are transformed into the local coordinate system of $\mathcal{G}_1$ using the transformation estimation result. A table in $\mathcal{G}_1$ is misclassified as a sofa (highlighted by the black rectangle), yet the transformation estimation remains correct and serves as a sufficient initial guess.}
\label{fig:map_semantic}\vspace{-5mm}
\end{figure}
\subsection*{ \hspace{1mm} A. Evaluation of Object Matching on Synthetic Data}

To quantitatively evaluate the robustness of the scene graph-based transformation estimation module, we create a $60 \text{m} \times 60 \text{m}$ environment containing 50 objects, each assigned a label from 6 distinct classes, along with randomly generated 
poses and dimensions.
This experiment simulates a hypothetical matching between robots, which can be extended to any n-robot scenario.
Tab.~\ref{tab:simulation_1} and~\ref{tab:simulation_2} show the success rate of transformation estimation under various conditions. 
Each experiment is repeated 20 times, with random initial transformations applied to both graphs for consistent results. 
The transformation estimation is considered successful if the translation error $e_t < 2 \text{m}$, and the rotation error $e_r < 20^\circ$ on each axis.
The recall represents the aggregate recall over all object matching results.
In both tables, the ``No Pose Error" column indicates when the true object 
pose is used for graph matching and transformation estimation, whereas the ``Pose Error" column corresponds to the case where the object pose includes random errors following a uniform distribution: $\pm1 \text{m}$ on the x- and y-axes, $\pm0.5 \text{m}$ on the z-axis, $\pm2.5 ^\circ$ for roll and pitch, and $\pm10^\circ$ for yaw.
While the runtime increases as the graph size grows, the proposed algorithm maintains real-time performance.
\begin{table}[htp]
\caption{\textbf{Success Rate} of transformation estimation and
\textbf{recall} of 
object matching for different graph sizes
and overlap levels, without semantic label errors.}
\vspace{-2mm}
\renewcommand{\arraystretch}{1.2}
\setlength{\tabcolsep}{3pt}
\label{tab:simulation_1}
\centering
\resizebox{0.97\columnwidth}{!}{%
\begin{tabular}{@{}ccc@{\hskip 5pt}cc@{\hskip 5pt}cc@{\hskip 5pt}c@{}}
\Xhline{2\arrayrulewidth}
\multicolumn{3}{c}{\textbf{Object Count}} &
\multicolumn{2}{c}{\textbf{Success Rate (\%)}} &
\multicolumn{2}{c}{\textbf{Recall (\%)}} &
{\textbf{Runtime (ms)}}\\
\scriptsize $\mathcal{G}_1$ &
\scriptsize $\mathcal{G}_2$ &
\scriptsize Overlap &
\scriptsize No Pose Err. &
\scriptsize Pose Err. &
\scriptsize No Pose Err. &
\scriptsize Pose Err. &
\scriptsize Time \\
\Xhline{2\arrayrulewidth}
8 & 8 & 4 & 100 & 65 & 100 & 100 & 7 \\
15 & 15 & 7 & 100 & 100 & 100 & 100 & 14 \\
35 & 32 & 17 & 100 & 100 & 100 & 100 & 226 \\
\Xhline{2\arrayrulewidth}
\end{tabular}%
}
\vspace{-2mm}
\end{table}
Tab.~\ref{tab:simulation_1} presents the success rate and recall for different graph sizes. 
Object count denotes the number of objects in each object graph, while overlap refers to the number of overlapping objects between two graphs. 
Our algorithm succeeds in all cases when no pose error is introduced. 
When pose error is present, it still achieves a relatively high success rate given a sufficient number of objects in the graph. 
However, the success rate is significantly lower in the 4-object overlap case, which is challenging due to the limited number of correct matches for transformation estimation.

\begin{table}[htp]
\caption{\textbf{Success Rate} of transformation estimation and \textbf{recall} of 
object matching with different semantic label errors.}\vspace{-3mm}
\renewcommand{\arraystretch}{1.2}
\label{tab:simulation_2}
\begin{center}
\resizebox{0.97\columnwidth}{!}{
\begin{tabular}{@{\hskip 7pt}c@{\hskip 6pt}cc@{\hskip 6pt}cc@{\hskip 6pt}c@{}}
\Xhline{2\arrayrulewidth}
\textbf{Mislabel} & 
\multicolumn{2}{c}{\textbf{Success Rate (\%)}} & 
\multicolumn{2}{c}{\textbf{Recall (\%)}} & {\textbf{Runtime (ms)}}\\ 
\textbf{Count} & \scriptsize No Pose Err. & \scriptsize Pose Err. & \scriptsize No Pose Err. & \scriptsize Pose Err. & \scriptsize Time\\ 
\Xhline{2\arrayrulewidth}
 3 & 100 & 100 & 82 & 82 & 209\\ 
 6 & 100 & 100 & 63 & 59 & 188\\ 
 9 & 95 & 95 & 46 & 46 & 183 \\  
12 & 55 & 30 & 29 & 27 & 178\\
\Xhline{2\arrayrulewidth}
\end{tabular}}
\end{center}\vspace{-5mm}
\end{table}

Tab.~\ref{tab:simulation_2} presents the success rate and recall when incorrect semantic classification is also considered.  
In all cases shown in Tab.~\ref{tab:simulation_2}, $\mathcal{G}_1$ contains 35 objects, and $\mathcal{G}_2$ contains 32 objects, with an overlap of 17 objects.  
Fig.~\ref{fig:map_semantic} illustrates a scenario where both pose and semantic errors are present, yet the proposed algorithm successfully handles the matching.  

\subsection*{B. SLAM Results over Real-World Data}
\begin{figure}
  \centering
    \subfigure{
        \includegraphics[width=.47\columnwidth]{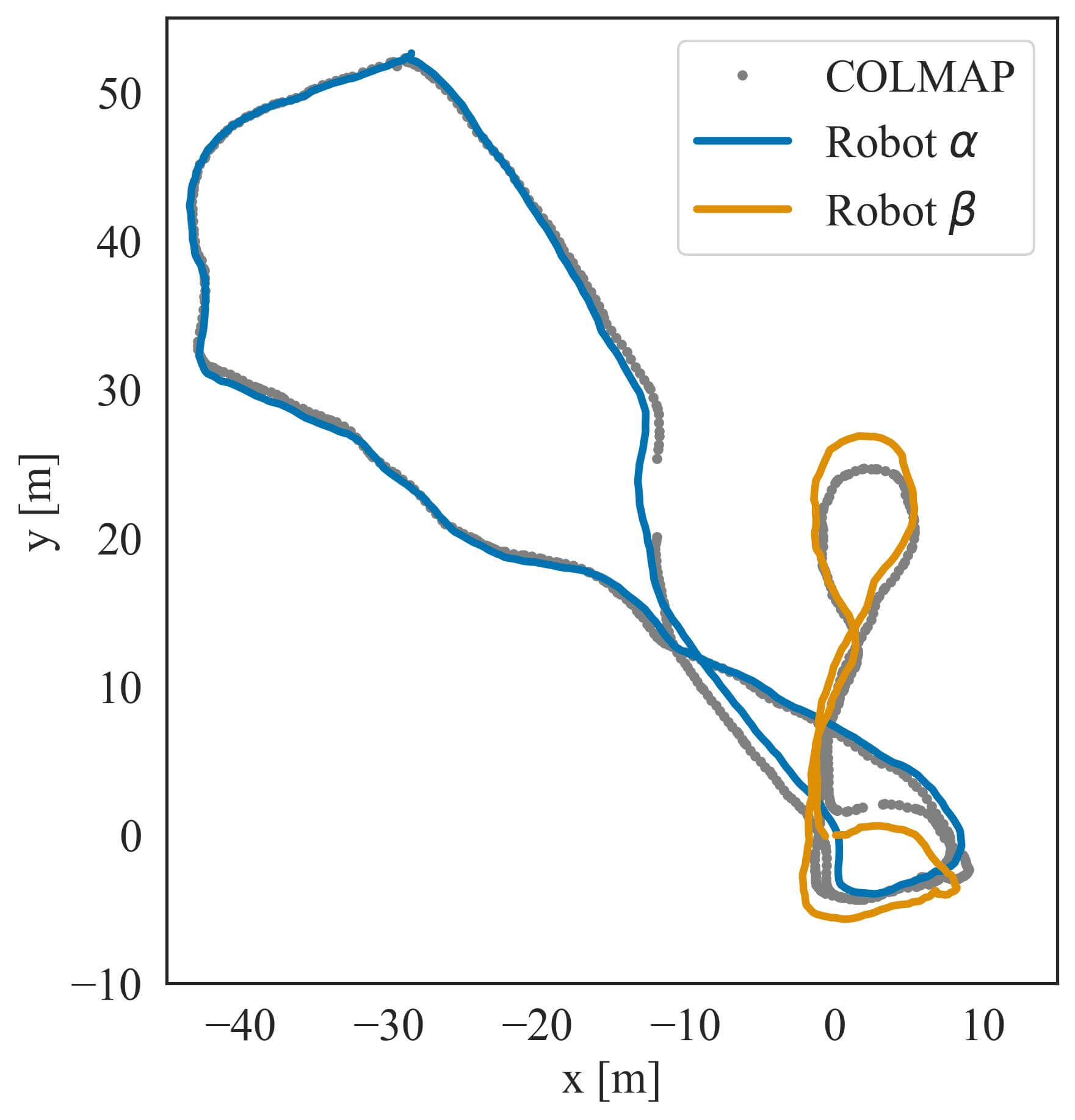}
    } 
    \subfigure{
        \includegraphics[width=.47\columnwidth]{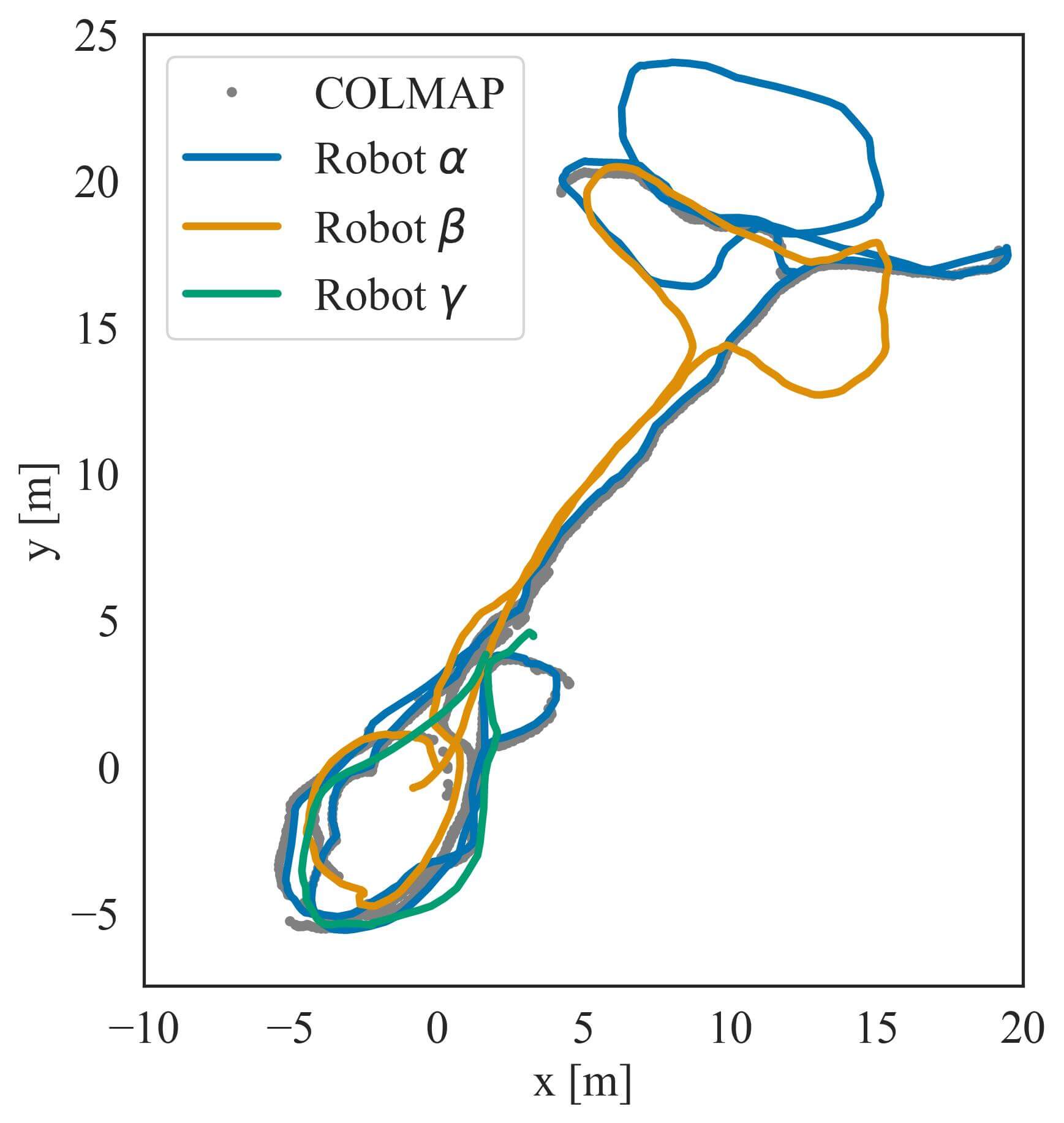}
    } 
  \caption{SLAM trajectories from multiple robots aligned with pose estimates from COLMAP sparse reconstruction, shown for our outdoor (left) and indoor (right) datasets.} \label{fig:traj}\vspace{-5mm}
\end{figure}

We compared our SGM-SLAM pipeline with a state-of-the-art LiDAR SLAM method on our self-gathered datasets.
Due to equipment limitations, GPS was not available for either the indoor or outdoor dataset. 
Instead, we used Structure from Motion (SfM) from COLMAP \cite{schonberger2016structure} as the ground truth, since it is a global method that optimizes observations across all images and performs inter-image registration globally.
The SfM results ensure accurate inter-robot relative pose estimation, which is crucial for evaluating multi-robot SLAM algorithms.
Swarm-SLAM is selected for comparison as it is a ROS2-based distributed method that supports both camera and LiDAR inputs. 
Its odometry is derived from LiDAR scans using RTAB-Map~\cite{labbe2019rtab}. 
Tab. \ref{tab:ate} shows the Root Mean Square Error (RMSE) of the Absolute Trajectory Error (ATE) between our SLAM and SfM results, while Fig. \ref{fig:traj} provides a qualitative comparison.
``Swarm-SC" denotes the use of the LiDAR descriptor Scan Context \cite{kim2018scan} for inter-robot data association, while ``Swarm-CP" denotes the use of the image descriptor CosPlace~\cite{Berton_CVPR_2022_CosPlace}.


We collected two datasets in the suburban campus environment.
The outdoor dataset consists of two robots covering a larger area and contains sparse and repeated human-made objects like chairs and benches, with some of the scenes occluded by trees, while the indoor dataset involves three robots covering a relatively small area with a large number of repeated and occluded objects like chairs, tables and bookshelves.

\begin{figure}
  \centering
    \subfigure{
        \includegraphics[width=.75\columnwidth]{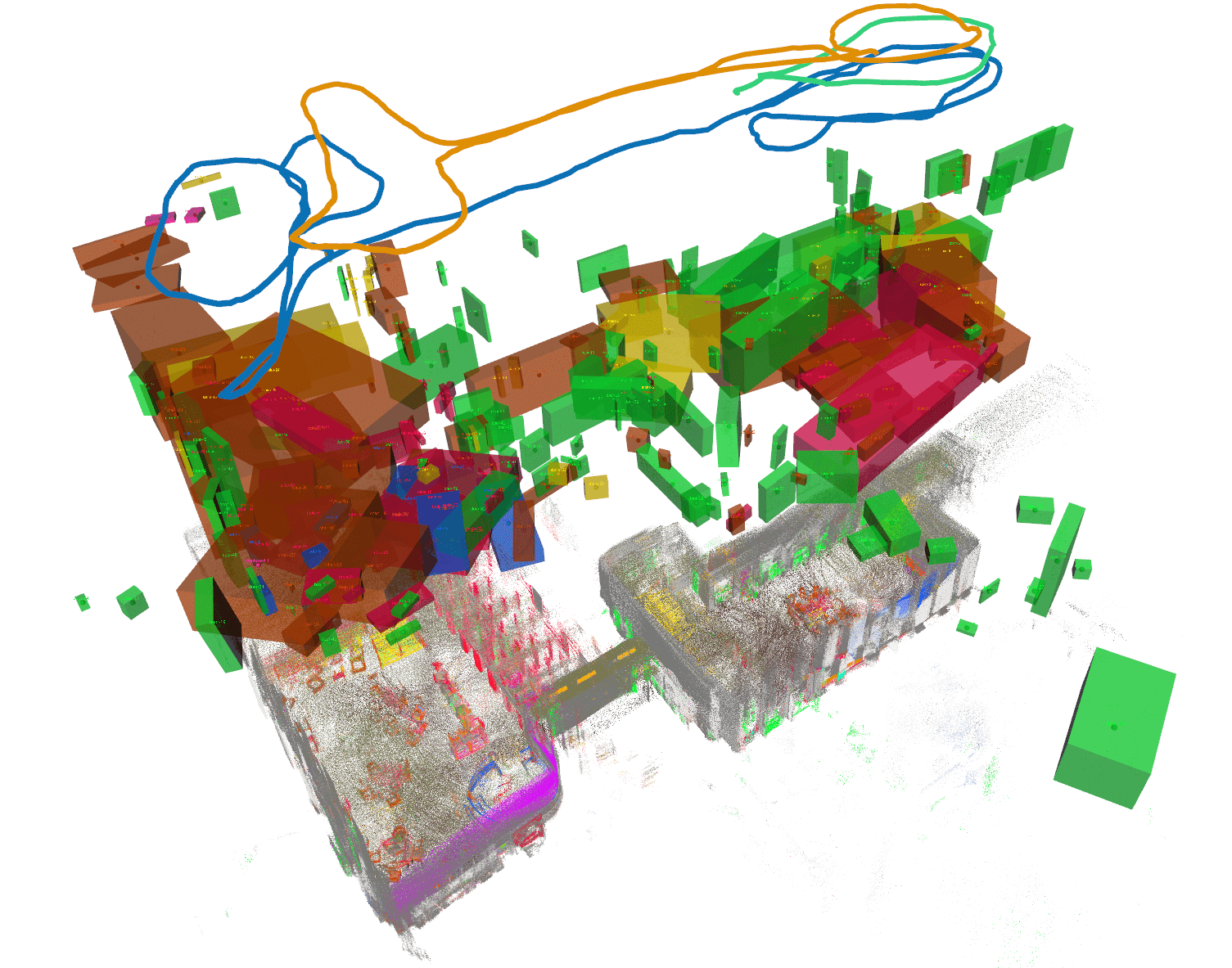}
    }
  \caption{The merged result from the three-robot dataset collected in an indoor area of the campus environment.} \label{fig:cloud_indoor}
  \vspace{-5mm}
\end{figure}

We use COLMAP to reconstruct only feature-rich areas, ensuring a precise reconstruction result as ground truth.
{\modified{The COLMAP reconstruction is aligned with the experimental result using a Sim(3) transformation.}}
The main challenge in the outdoor dataset is the low overlap ratio between trajectories, requiring robust inter-robot data association. 
The indoor dataset, on the other hand, is affected by dim lighting conditions. 
Additionally, both datasets contain repeated feature patterns from objects such as chairs, benches, and tables.
To align the trajectories and evaluate error, we use EVO\footnote{\vspace{-5mm}https://github.com/MichaelGrupp/evo}.
Since EVO is designed for single-robot usage, we combine the trajectories collected by different robots into a single file and evaluate the results as a whole trajectory.
The merged result based on scene graph matching from the outdoor dataset is shown in Fig.~\ref{fig:map_scene}, while the result from the indoor dataset is shown in Fig.~\ref{fig:cloud_indoor}. 
Swarm-SC performs poorly on the outdoor dataset because the two trajectories share only a small portion of overlap, whereas Scan-Context is effective primarily when there is a large enough viewpoint overlap.
Swarm-CP does not perform well in the indoor environment due to dim lighting and the difference in viewpoint height between the human operator and the robot platform.
However, because our object-based scene graph matching strategy treats the entire object graph as a single entity, it is more robust to viewpoint changes and limited trajectory overlap.

\vspace{-2mm}
\begin{table}[htp]
\caption{\textbf{Absolute Trajectory Error (ATE) [m]} in meters for the proposed method compared to state-of-art methods.} 
\vspace{-2mm}
\renewcommand{\arraystretch}{1.2}
\label{tab:ate}
\begin{center}
\begin{tabular}{@{\hskip 13pt}c@{\hskip 13pt}c@{\hskip 13pt}c@{\hskip 13pt}c@{\hskip 13pt}c@{\hskip 13pt}}
\Xhline{2\arrayrulewidth}
\textbf{Method} & {\textbf{Swarm-SC}} & \textbf{Swarm-CP} & {\textbf{Ours}}\\ 
\Xhline{2\arrayrulewidth}
 Outdoor  & 20.29 & 3.92 & 1.32 \\ 
 Indoor   & 3.36 & 5.70 & 0.81\\
\Xhline{2\arrayrulewidth}
\end{tabular}
\end{center}\vspace{-5mm}
\end{table}
\vspace{-3mm}
\subsection*{C. Communication Analysis}
We next analyze the size and number of perception messages exchanged in our real-world datasets using simulated wireless communication. 
We summarize the size of perception messages for the proposed method in Tab.~\ref{tab:message}.
The ``Point-cloud Raw'' baseline we compared against refers to a brute-force centralized strategy, where all raw point clouds are transmitted to a centralized server.
To further reduce the communication overhead of our framework, only limited object information is shared, which includes the object ID, centroid position, dimensions and semantic label. 
Object data is only exchanged when there is a significant update to the scene graph, such as when a new object is added or a large odometry drift is corrected.
This strategy enables us to publish the object data at a much lower frequency compared to the LiDAR sensor frequency.

Once a relative graph transformation estimation is computed, the relevant keyframes are requested from the robot neighbors. 
A keyframe is only sent after the requesting robot confirms it has not been sent previously.
The point cloud keyframe message is compressed using PCL~\cite{Rusu_ICRA2011_PCL} to minimize its size.
During the experiments, we observed that keyframe requests were sent at a much higher frequency than the keyframe responses. 
This occurred mainly because, once two scene graphs are matched, a burst of similar requests are sent to the neighbor, causing the oldest requests to be dropped.
Further optimization, such as implementing a delayed request strategy, may be needed to improve performance further.
\vspace{-2mm}
\begin{table}[htp]
\caption{\textbf{Sizes of Perception Messages [KB]} for the
proposed method. ``No.'' indicates the number of messages published during a dataset run.}\vspace{-4mm}
\renewcommand{\arraystretch}{1.2}
\label{tab:message}
\begin{center}
\begin{tabular}{@{\hskip 9pt}c@{\hskip 8pt}ccc@{\hskip 8pt}ccc@{\hskip 8pt}}
\Xhline{2\arrayrulewidth}
\multirow{2}{*}{\textbf{Message Type}} & 
\multicolumn{3}{c}{\textbf{Outdoor}} & \multicolumn{3}{c}{\textbf{Indoor}}\\ 
 & Mean & Max & No.  & Mean & Max & No. \\ 
\Xhline{2\arrayrulewidth}
 Point-cloud Raw & 508.2& 516.9 & 3073 & 508.1 & 512.1 & 3079\\
\Xhline{1\arrayrulewidth}
 Object Data & 7.76  & 16.93  & 701 & 18.40  & 44.37 & 845\\ 
 Keyframe Request & 0.02 & 0.02  & 61 & 0.02 & 0.02 & 411\\
 Keyframe Response & 18.87 & 22.95 & 10 & 6.90 & 11.05 & 72\\ 
\Xhline{2\arrayrulewidth}
\end{tabular}
\end{center}\vspace{-5mm}
\end{table}
\vspace{-3mm}
\section{Conclusion}\label{sec:conclusion}
In this paper, we propose SGM-SLAM, a data-efficient distributed SLAM method using limited object data scene graph matching.
We extract geometric and semantic information from LiDAR and camera data, and construct a multi-level scene graph collaboratively with the multi-robot team. 
The introduction of object-level graph matching improves the generality, robustness, and communication efficiency of our distributed SLAM system in both indoor and outdoor environments.
\modified{Future work will evaluate the framework under explicitly bandwidth-limited and intermittent communication conditions and investigate deployment in larger-scale field environments.}






\vspace{-3mm}
\bibliographystyle{IEEEtran}
\bibliography{bib_abb}

\end{document}